%% file: main.tex
\documentclass[runningheads]{llncs}

\usepackage{eccv}

\usepackage{eccvabbrv}

\usepackage{graphicx}
\usepackage{booktabs}
\usepackage{algorithm}
\usepackage{algorithmic}

\usepackage{amsmath, bm}
\usepackage{amssymb}
\usepackage{booktabs}
\usepackage{multicol}
\usepackage{multirow}
\usepackage{makecell}
\usepackage{colortbl}

\usepackage[accsupp]{axessibility}  

\usepackage[pagebackref,breaklinks,colorlinks,citecolor=eccvblue]{hyperref}

\usepackage{orcidlink}

\begin{document}

\title{Online Open-set Semi-supervised Object Detection with Dual Competing Head} 

\titlerunning{Abbreviated paper title}

\author{
    Zerun Wang$^1$, Ling Xiao$^{1}$,Liuyu Xiang$^2$, \\Zhaotian Weng$^3$, Toshihiko Yamasaki$^1$
}

\authorrunning{F.~Author et al.}

\institute{The University of Tokyo \and
Beijing University of Posts and Telecommunications \and
Tsinghua University\\
\email{\{ze\_wang,ling,yamasaki\}@cvm.t.u-tokyo.ac.jp} \\
\email{\{xiangly\}@bupt.edu.cn}
\email{\{zhaotianweng\}@gmail.com}
}

\maketitle

\begin{abstract}
Open-set semi-supervised object detection (OSSOD) task leverages practical open-set unlabeled datasets that comprise both in-distribution (ID) and out-of-distribution (OOD) instances for conducting semi-supervised object detection (SSOD). The main challenge in OSSOD is distinguishing and filtering the OOD instances (i.e., outliers) during pseudo-labeling since OODs will affect the performance. The only OSSOD work employs an additional offline OOD detection network trained solely with labeled data to solve this problem. However, the limited labeled data restricts the potential for improvement. Meanwhile, the offline strategy results in low efficiency. To alleviate these issues, this paper proposes an end-to-end online OSSOD framework that improves performance and efficiency: 1) We propose a semi-supervised outlier filtering method that more effectively filters the OOD instances using both labeled and unlabeled data. 2) We propose a threshold-free Dual Competing OOD head that further improves the performance by suppressing the error accumulation during semi-supervised outlier filtering. 3) Our proposed method is an online end-to-end trainable OSSOD framework. Experimental results show that our method achieves state-of-the-art performance on several OSSOD benchmarks compared to existing methods. Moreover, additional experiments show that our method is more efficient and can be easily applied to different SSOD frameworks to boost their performance.
  \keywords{Object detection \and Semi-supervised learning \and Open-set problem}
\end{abstract}

\section{Introduction}
\label{sec:intro}

Semi-supervised learning (SSL) significantly improves the performance of various image recognition tasks by utilizing a large amount of available unlabeled data~\cite{sohn2020fixmatch,berthelot2019mixmatch,sohn2020simple,tarvainen2017mean,yu2020multi}. Object detection performance has also greatly benefited from SSL, leading to the proposal of various semi-supervised object detection (SSOD) methods~\cite{liuunbiased,li2022dtg,li2022pseco,xu2021end,wang2023consistent,zhou2022dense}. 

\begin{figure}[t]
  \centering
  \includegraphics[width=1.0\linewidth]{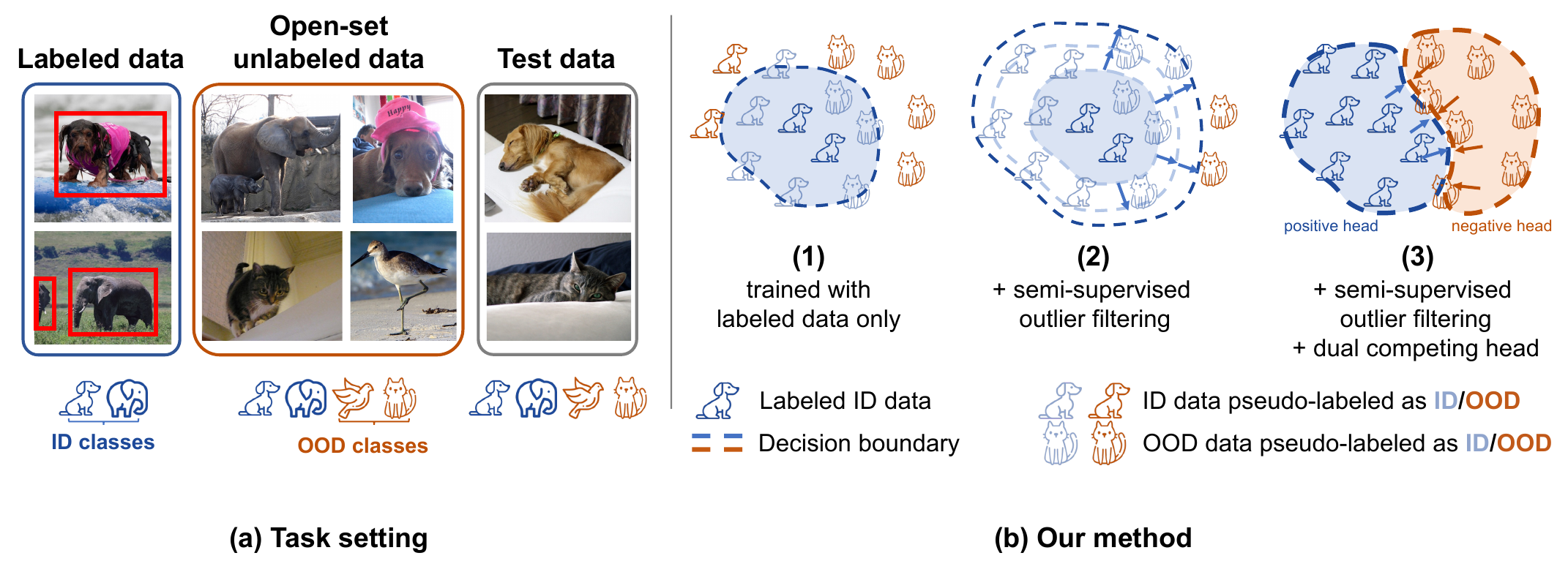}
  \caption{\textbf{(a)} The data setting of the OSSOD task. \textbf{(b)} 1) The previous OSSOD method trained the model with only labeled data. 2) We first improve the performance by our semi-supervised outlier filtering method but face the error accumulation problem: The mispredicted OODs make the decision boundary expand to misclassify more samples. 3) We further propose the Dual Competing OOD head to alleviate the error accumulation and result in better performance.}
  \label{fig:compare}
\end{figure}

However, current SSOD methods are under a strong assumption that the unlabeled and labeled data are from the same label space. This assumption is somewhat unrealistic in practical situations because the unlabeled dataset in the real world usually faces the open-set problem, which means there are OOD samples as shown in Fig.~\ref{fig:compare}(a). Object detectors in current SSOD methods will mistakenly classify OOD samples as ID classes and ultimately degrade the performance.

Some works~\cite{yu2020multi,saito2021openmatch} have been proposed to tackle this problem in the image classification task. However, these methods are difficult to apply to object detection tasks directly since image classification is an image-level task, but object detection is a more challenging instance-level task.  
The paper of~\cite{liu2022open} is the first to tackle the open-set problem in object detection and name it as the OSSOD task. They tried to apply existing OOD detection methods directly but found that their performance was not satisfactory. Then they proposed the first OSSOD method by training a separate OOD detection network with labeled data to distinguish and filter out OOD instances for the SSOD framework during pseudo-labeling. 

Although they have improved the performance on open-set unlabeled data, there are still some challenges that need to be addressed: First, they only use labeled data to train the OOD detection network. However, in the OSSOD task, real OOD instances only exist in unlabeled data. The lack of real OODs results in suboptimal performance, as shown in Fig.~\ref{fig:compare}(b-1). Second, they need a manual threshold for the OOD detection network to filter out OOD instances. It is time-consuming to search for the best threshold for each dataset. Third, their OOD detection network needs to be trained separately from the object detector and requires an additional backbone, which is inefficient considering the training process and network complexity. 

To address the above issues, we propose a novel OSSOD method: 1) We propose a semi-supervised outlier filtering strategy to improve OOD filtering ability by leveraging both labeled and unlabeled data. 2) We further identify the error accumulation problem: the mispredictions in pseudo-labels accumulate during semi-supervised outlier filtering. As shown in Fig.~\ref{fig:compare}(b-2), once the OOD instances are mispredicted as ID (the blue cats), the decision boundary expands to misclassify more OOD labels. To tackle this, we propose the Dual Competing OOD~(DCO) head, which mitigates this issue with two sub-heads that form a competitive relationship during semi-supervised learning as shown in Fig.~\ref{fig:compare}(b-3) and further improves the performance. 
Meanwhile, the DCO head does not require any manual threshold for filtring OOD instances. 3) We render the entire OSSOD framework online end-to-end trainable. 

The experimental results on several benchmarks show that our method can achieve state-of-the-art OSSOD performance. Meanwhile, our method can be easily applied to other SSOD frameworks to boost their performance. In summary, this paper presents the following contributions:

\begin{itemize}
    \item We propose a semi-supervised outlier filtering strategy, which improves the OSSOD accuracy by better utilizing the unlabeled data.
    
    \item We further identify and mitigate the error accumulation problem in semi-supervised outlier filtering by the threshold-free Dual Competing OOD head. 
    
    \item The above two components constitute an online end-to-end OSSOD framework. Our proposed method achieves state-of-the-art performance on several OSSOD benchmarks and can be applied to other SSOD frameworks. 
    
\end{itemize}

\section{Related Work}

\subsection{Semi-supervised object detection} 
Semi-Supervised Object Detection (SSOD) methods aim to improve object detection performance 
with unlabeled data. Some basic SSOD technologies are transferred from semi-supervised image classification tasks such as data augmentation~\cite{berthelot2019mixmatch}, teacher-student framework~\cite{sohn2020simple}, and exponential moving average (EMA)~\cite{tarvainen2017mean}. Recent SSOD research addresses unique object detection problems, such as class-wise imbalance~\cite{liuunbiased}, localization reliability~\cite{li2022rethinking,xu2021end,li2022pseco}, dynamic thresholding~\cite{wang2023consistent}, and using dense learnable regions over hard pseudo-labels~\cite{zhou2022dense}. However, these methods neglect the presence of OOD instances in open-set unlabeled data. It has been shown that pseudo-labels containing OODs lead to the semantic expansion problem and affect the performance~\cite{liu2022open}.

\subsection{Open-set semi-supervised learning} 
Most of the open-set semi-supervised learning (OSSL) methods~\cite{li2023iomatch,he2022safe} focus on image classification tasks. Yu et al.~\cite{yu2020multi} proposed a multi-task curriculum learning method to select ID samples from unlabeled data by alternatively estimating the OOD score for unlabeled images and training the network. Saito et al. ~\cite{saito2021openmatch} relies on the one-vs-all OOD detection method to filter OOD samples after pseudo-labeling and use a consistency regularization loss to learn more effective representations. However, these methods are incompatible with object detection tasks: The main difference is that each image contains one object in the image classification task but contains a variable number of objects in the object detection task. Moreover, the number of detected objects in each image is also variable during training. As a result, we cannot maintain a fixed number of OOD scores as~\cite{yu2020multi} or augment the image several times for each object to compute the consistency regularization loss such as~\cite{saito2021openmatch} considering the complexity.  OSSL methods also take some techniques from the OOD detection task~\cite{hendrycks2016baseline, lee2018simple, liu2020energy, du2022vos}. However, OOD detection aims to train on a large number of labeled ID data to distinguish OOD samples, which is different from the OSSL setting.

Liu et al. ~\cite{liu2022open} proposed the only work of OSSL on the object detection task: the outliers are filtered by a pre-trained OOD detection network. However, the OOD detection network is trained separately and only with labeled data. We further improved the accuracy and efficiency.


\subsection{Open-set object detection} 
The open-set object detection (OSOD) task focuses on detecting both ID and unknown OOD objects. Early approaches use dropout sampling~\cite{miller2018dropout} to reduce open-set errors. OWOD~\cite{joseph2021towards} utilizes the energy score to discern known and unknown classes. OpenDet~\cite{han2022opendet} separates ID and OOD samples by identifying high/low-density regions in the latent space. The OSOD task is different from the OSSOD task in that OSOD  seeks to enhance the accuracy of both ID and OOD classes, while OSSOD focuses on the performance of ID classes and prevents the detrimental effects caused by distracting OOD objects. Meanwhile, these methods also rely on substantial labeled data for training. 

\section{Preliminary}

OSSOD task aims to solve the open-set problem in SSOD. Thus, OSSOD methods are based on SSOD frameworks. The SSOD task assumes that the object detector is trained on both labeled dataset $\bm{D}_l=\{\bm{X}_l, \bm{Y}_l\}$ and unlabeled dataset $\bm{D}_u=\{\bm{X}_u\}$. A common pipeline is setting two detectors: student and teacher models. The teacher model generates pseudo-labels $\bm{\hat{Y}}_u$ for unlabeled data $\bm{D}_u$. The generated pseudo-labels are then selected by a manually set threshold $\tau$ on the classification confidence score. Then, the student model is jointly trained with labeled data and pseudo-labeled unlabeled data. The total loss function is defined as: 
\begin{align}
\label{eq:ssodloss}
    \mathcal{L}^{ssod} = \mathcal{L}_{sup}^{ssod}(\bm{X}_l, \bm{Y}_l) + \lambda\mathcal{L}_{unsup}^{ssod}(\bm{X}_u, \delta(\bm{\hat{Y}}_u, \tau)),
\end{align}
where $\mathcal{L}_{sup}$ and $\mathcal{L}_{unsup}$ denote the loss function for training with labeled data and unlabeled data, respectively. Each consists of classification and regression losses in object detection tasks. $\lambda$ controls the weight of learning with unlabeled data and $\delta(\cdot)$ denotes the thresholding process. During training, the teacher model is updated by the student's parameters using the exponential moving average (EMA) method to get a more stable result.

\begin{figure*}[t]
  \centering
  \includegraphics[width=1.0\linewidth]{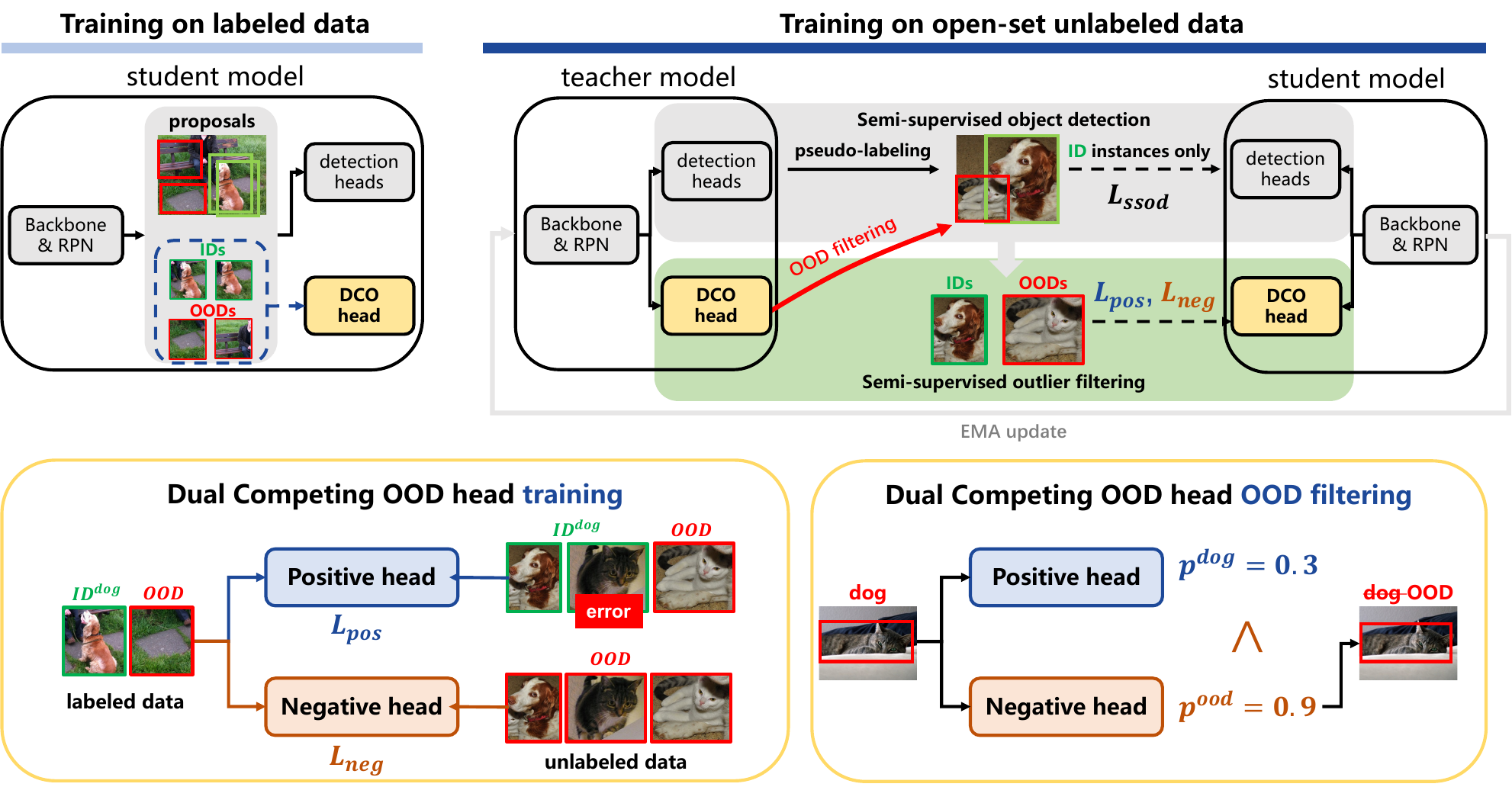}
  \caption{The framework of our method. \textbf{Top:} Our DCO head is added to the detector for filtering OODs in the pseudo-labels during training. We propose the semi-supervised outlier filtering strategy to improve the filtering ability. \textbf{Bottom-left:} Training strategy of our DCO head, the pseudo-labeled ID/OODs are used for training the positive head (Note that wrong pseudo-label exists). We label all the unlabeled instances as OOD for training the negative head. \textbf{Bottom-right:} OOD filtering using the DCO head. Two heads compete with each other to decide on ID or OOD. In this case, \textit{dog} is the ID class, and \textit{cat} is the OOD class.}
  \label{fig:framework}
\end{figure*}

\section{Method}

We apply the Unbiased Teacher~\cite{liuunbiased}, which follows the preliminary, as our baseline SSOD method. Fig.~\ref{fig:framework} illustrates the entire structure of our framework: Our proposed Dual Competing OOD (DCO) head is added to the object detector to filter the OOD instances in the pseudo-labels for SSOD. The DCO head is trained with both labeled and unlabeled data using our semi-supervised outlier filtering strategy. Our framework is online end-to-end trainable. In this section, we first introduce the semi-supervised filtering strategy. Then, we introduce the details of our DCO head.

\subsection{Semi-supervised outlier filtering}

The previous method~\cite{liu2022open} trains the outlier filtering network with labeled data only. However, the real OOD instances only exist in the unlabeled data in the OSSOD setting. Thus, we aim to further utilize the unlabeled data to improve the filtering ability. To achieve this, we introduce an OOD detection head into the object detector to filter OOD instances (We can use either previous OOD detection head structures or our DCO head). The head takes the feature of proposals after ROI-pooling as input and predicts the probability of each sample belonging to ID or OOD classes. We train the head with both labeled and unlabeled data in a semi-supervised way.

\textbf{Training on labeled data.} Since the labeled data provide reliable supervision, our OOD detection head also relies on training with the annotations from labeled data. Following~\cite{liu2022open}, we use the proposals from the RPN network with high overlap to the ground-truth as ID instances and those proposals with low overlap as OOD instances to train the OOD detection head. The overlap threshold here is consistent with the one used in distinguishing foreground and background in the original object detection task. For each image, we collect a fixed number of instances to form a batch: we first collect all the ID instances and then randomly gather OOD instances until the batch size is complete.

\textbf{Training on unlabeled data.} When training on the unlabeled data, we first get the pseudo-labeled instances from the original detection heads and label them as ID or OOD regarding the prediction of our OOD detection head. Then we use these instances to train the student's OOD detection head. It is worth noting that we can get real OOD instances and more ID instances from unlabeled data to train the head in this way. Thus, the OOD head can be exposed to a broader range of distribution characteristics present in the unlabeled data, thereby improving the performance. The parameters of the OOD detection head are updated using the EMA during SSOD, thus, our method also benefits from the stable ID and OOD predictions from the teacher model's OOD detection head. To make the training stable, we also sample background proposals from unlabeled data to maintain the fixed batch size mentioned above.

\subsection{Dual Competing OOD head}

\begin{figure}[t]
  \centering
  \includegraphics[width=0.8\linewidth]{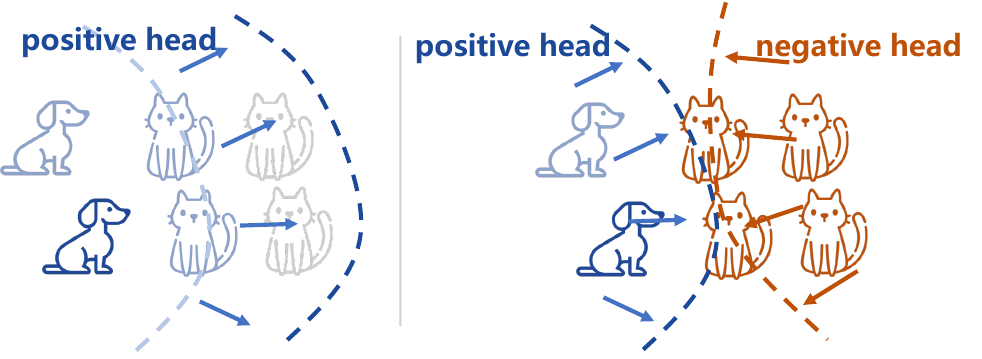}
  \caption{Left: The error accumulation problem with only one OOD detection head. Right: The principle of our DCO head for preventing the problem. In this case, \textit{dog} is the ID class, and \textit{cat} is the OOD class. The dashed line represents the decision boundary.}
  \label{fig:analyze}
\end{figure}

We experimentally find that two problems arise during semi-supervised outlier filtering when we directly apply the OOD detection head structure in the pioneering OSSOD method\cite{liu2022open}: 1) As shown in Fig.~\ref{fig:analyze}, the OOD detection head will inevitably generate incorrect predictions, such as labeling OOD instances as ID. If such pseudo-labels are used for semi-supervised outlier filtering, the model will gradually accumulate more errors. 2) A threshold for distinguishing ID and OOD instances is needed for the previous OOD detection head. And it is time-consuming to find the proper threshold for different datasets. We propose the DCO head to solve these problems and further improve the performance.



\textbf{Analysis of error accumulation problem.} Applying the previous OOD detection head will cause the error accumulation problem because there is no mechanism to recheck whether a prediction is correct once it has a confidence score above the threshold. Therefore, we aim to add an additional module to constrain the original OOD detection head during the entire training process as shown in Fig.~\ref{fig:analyze}.

\textbf{DCO head structure.} Our DCO head consists of two sub-classifiers: the positive head is used for our proposed semi-supervised outlier filtering, while the negative head is used for constraining the positive head. The two heads are both $K+1$ classifiers ($K$ ID classes and one OOD class) with the same structure. When determining whether a sample is ID or OOD, the two heads form a competitive relationship:
Suppose we have an instance $\bm{x}$ with class prediction $y$ from the object detector's classification head. It will be determined as ID only when its confidence score of the $y^{th}$ class in the positive head $p_{pos}^{y}$ surpasses the confidence score of the OOD class in the negative head $p_{neg}^{K+1}$:
\begin{align}
\bm{x} \  {\rm is} \ \begin{cases}
\mathbf{id}, & {\rm if}\  p_{pos}^{y}\geq\ p_{neg}^{K+1}, \\
\mathbf{ood}, & {\rm otherwise}.  \\
\end{cases}
\end{align}
With this structure, no additional threshold is needed to filter OOD instances. 

\begin{algorithm}[t]
\caption{Our Online OSSOD method}
\label{alg:alg1}
\textbf{Input}: Labeled data: $\bm{D}_l=\{\bm{X}_l, \bm{Y}_l\}$, Unlabeled data: $\bm{D}_u=\{\bm{X}_u\}$\\
\textbf{Output}: {Parameters of teacher and student model $\theta_t,\theta_s$}.

\begin{algorithmic}[1] 
\FOR {burn-in iterations}
\STATE Compute SSOD loss $\mathcal{L}^{ssod}$ on labeled data.
\STATE Compute OOD detection loss $\mathcal{L}^{DCO}$ on labeled data.
\STATE Update student model $\theta_s =\theta_s - \nabla\mathcal{L}$.
\ENDFOR
\STATE initiate teacher model $\theta_t = \theta_s$ \;
\FOR {semi-supervised learning iterations}
\STATE Generate pseudo-labels for unlabeled data: $\bm{Y}_u=f(\bm{X}_u;\theta_t)$.
\STATE Apply SSOD thresholding for pseudo-labels: $\hat{\bm{Y}}_u = \delta(\bm{Y}_u; \tau)$.
\STATE Distinguish the pseudo-labels as ID and OOD classes by the DCO head: $\hat{\bm{Y}}_u = \{\hat{\bm{Y}}_u^{id}, \hat{\bm{Y}}_u^{ood}\}$.
\STATE Compute SSOD loss $\mathcal{L}^{ssod}$ on labeled data $\{\bm{X}_l, \bm{Y}_l\}$, and unlabeled ID data $\{\bm{X}_u, \hat{\bm{Y}}_u^{id}\}$.
\STATE Compute OOD detection loss $\mathcal{L}^{DCO}$ on labeled data $\{\bm{X}_l, \bm{Y}_l\}$, and unlabeled data $\{\bm{X}_u, \{\hat{\bm{Y}}_u^{id},\hat{\bm{Y}}_u^{ood}\}\}$.
\STATE Jointly train student model with the losses $\mathcal{L} = \mathcal{L}^{ssod} + \lambda^{ood}\mathcal{L}^{DCO}$.
\STATE Update student model $\theta_s \gets \theta_s - \nabla\mathcal{L}$.
\STATE Update teacher model by EMA $\theta_t \gets \alpha \theta_t + (1-\alpha)\theta_s$\;
\ENDFOR
\STATE \textbf{return} Parameters $\theta_t,\theta_s$ of teacher and student model
\end{algorithmic}
\end{algorithm}

\textbf{Competing training strategy.} We propose a competing training strategy for the DCO head. Specifically, both heads are trained with the cross-entropy loss. When training with labeled data, both the positive and the negative heads share the same label since the labeled data is reliable. When training with unlabeled data, the positive head will follow the semi-supervised learning scheme to use the pseudo-labels from its own prediction. However, the negative head will treat all the instances as OOD since they are not inherently reliable. The overall loss is as follows:
\begin{gather}
\mathcal{L}^{DCO} = \mathcal{L}_{pos} + \mathcal{L}_{neg}, \\
\mathcal{L}_{pos} = \frac{1}{N_l}\sum_{i=0}^{N_l}l_{cls}(\bm{x}_l^i, y_l^i)+\frac{1}{N_u}\sum_{i=0}^{N_u}l_{cls}(\bm{x}_u^i, y_u^i), \\
\mathcal{L}_{neg} = \frac{1}{N_l}\sum_{i=0}^{N_l}l_{cls}(\bm{x}_l^i, y_l^i)+\frac{1}{N_u}\sum_{i=0}^{N_u}l_{cls}(\bm{x}_u^i, y_{ood}^i), 
\end{gather}
where $l_{cls}$ denotes the cross-entropy loss, $\bm{x}_l, \bm{x}_u$ denotes the labeled and unlabeled instances in a single batch with batch size $N_l, N_u$, respectively. $y_l^i\in\{1,...,K+1\}$ is the provided label of $x_l$. $y_u^i\in\{1,...,K+1\}$ is the pseudo-label from the DCO head for $x_u$. $y_{ood}^i\in\{K+1\}$ is the OOD label for negative head.

With our DCO head, the negative head will have high OOD confidence scores for all pseudo-labels, especially for those unseen OOD objects that significantly differ from the ID instances. Therefore, even if the positive head mispredicts an OOD instance as an ID class, the negative head can still prevent this mistake since the corresponding OOD confidence score can also be high. The experimental results prove the effectiveness of our DCO head. 

We combine $\mathcal{L}^{DCO}$ with the loss function of our based SSOD framework to train the model:
\begin{align}
\label{eq:totalloss}
    \mathcal{L} = \mathcal{L}^{ssod} + \lambda^{ood}\mathcal{L}^{DCO},
\end{align},
where $\lambda^{ood}$ controls the weight of $\mathcal{L}^{DCO}$. Our entire training process can be described as in Alg.~\ref{alg:alg1}.

\section{Experiments}

\subsection{Datasets and evaluation metrics}
Our method is evaluated on the COCO-Open and COCO-OpenImages datasets proposed by the pioneering work of OSSOD~\cite{liu2022open}. We also evaluate our method on the newly introduced COCO-VOC dataset.

\textbf{COCO-Open.} We randomly select 20/40/60 classes as ID classes and the remaining as OOD classes in the MS-COCO 2017~\cite{lin2014microsoft} dataset with 80 classes. The training set is divided into ID, MIX, and OOD sets by splitting the classes. The images in the ID set contain only instances of ID classes. The images in the OOD set contain only instances of OOD classes. The images in the MIX set contain both instances of ID and OOD classes. We then randomly sample images with annotations from the ID set as the labeled dataset. The rest of the ID set and other sets are combined as the open-set unlabeled dataset. For evaluation, we use all the images in the MS-COCO 2017 validation set but delete the annotations of OOD instances. 

\textbf{COCO-OpenImages.} We also evaluate our method on a large-scale dataset, using the entire MS-COCO 2017 as the labeled ID dataset and OpenImagesv5~\cite{OpenImages2} as the open-set unlabeled dataset. OpenImagesv5 contains 1.7M images with 601 classes. Classes not present in MS-COCO are considered as OOD classes. For evaluation, we use the entire MS-COCO 2017 validation set.

\textbf{COCO-VOC.} The Pascal-VOC 2012 dataset~\cite{everingham2010pascal} consists of 20 classes, all of which fall within the 80 classes of the COCO dataset. We employ the Pascal VOC training set as our labeled data and the MS-COCO training set as our unlabeled data. For evaluation, we use both the MS-COCO and Pascal-VOC validation sets.

\textbf{Evaluation metrics.} We use the standard mean Average Precision (mAP) to evaluate the object detection performance and the area under the ROC curve
(AUROC) to evaluate the OOD detection performance. To calculate AUROC for object detection, we label all detection results as either ID or OOD classes, depending on whether their IoU score with the annotations (containing only ID instances) exceeds 0.5.

\subsection{Baseline methods}

We mainly compare our method with the first OSSOD work~\cite{liu2022open} (referred to as offline OSSOD for convenience). This work is based on the SSOD framework Unbiased Teacher (UT)~\cite{liuunbiased}. We also apply some OOD detection and open-set object detection methods for ablation studies, including OE~\cite{hendrycks2018deep}, Energy~\cite{liu2020energy}, OVA-Net~\cite{saito2021ovanet}, VOS~\cite{du2022vos}, and OpenDet~\cite{han2022opendet}.

\subsection{Implementation details}

For a fair comparison, we mainly use UT~\cite{liuunbiased} as the basic SSOD framework, which uses Faster R-CNN~\cite{ren2015faster} with Feature Pyramid Network~(FPN)~\cite{lin2017feature} and ResNet-50~\cite{he2016deep} backbone. We keep the same hyper-parameter settings with UT and offline OSSOD, including the learning rate, SSOD thresholds, training schedule, etc. The only new hyper-parameter of our work is the weight $\lambda^{ood}$ of the OOD detection loss $\mathcal{L}^{DCO}$. We set it to 0.1. The other hyper-parameters are reported in the appendix. The whole framework is based on Detectron2~\cite{wu2019detectron2}. 

\subsection{Experiments on OSSOD benchmarks}

\textbf{Varying number of ID classes and labeled images.} We evaluate our method by using various numbers of ID classes (20/40/60) and labeled images (1000/2000/4000). We run each experiment 3 times and report the standard deviation. The results in Table~\ref{tab:coco_class} and Table~\ref{tab:coco_label} show that our method consistently outperforms the offline OSSOD method across various settings. In half of the cases, our improvement based on UT is more than double that of the previous method. Details of the selected ID classes are provided in the appendix. Meanwhile, we find that as the number of ID classes increases to 60, the improvement of OSSOD methods tends to decrease. This can be attributed to that with the fixed total class number, when the number of ID classes increases, the model will acquire strong class-wise distinguishing abilities. And the impact of a small number of OOD classes naturally diminishes. Similarly, when the number of ID classes is small, our OSSOD method leads to more substantial improvement.

\begin{figure}[t]
  \centering
  \includegraphics[width=1.0\linewidth]{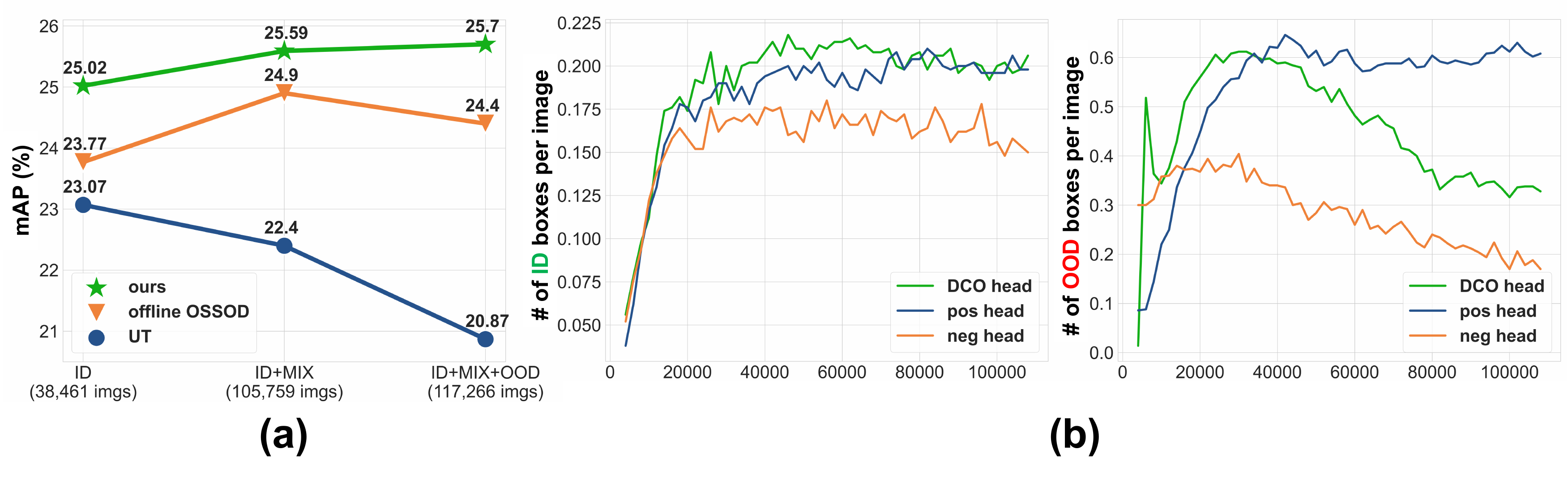}
  \caption{\textbf{(a)} Performance under different data combinations. \textbf{(b)} The number of ID (left) and OOD (right) pseudo-labeled boxes per image during training for different heads. Pos and neg denote our positive and negative heads respectively.}
  \label{fig:coco_id}
\end{figure}

\textbf{Effect of different unlabeled data combinations.}
We further show the effectiveness of our method using different combinations of unlabeled data. We use COCO-Open with 40 ID classes and 4000 labeled images. Then, we consider different unlabeled data combinations of ID, ID+MIX, and ID+MIX+OOD sets. The results in Fig.~\ref{fig:coco_id}(a) show that 1) we once again demonstrate that OOD samples are detrimental to the SSOD task, as the performance of UT continuously decreases when introducing more OOD instances, while OSSOD methods can alleviate this problem. 2) With the increase of OOD instances, the performance of the previous OSSOD method also declined, which suggests that our method is more robust. Meanwhile, although there is no ID foreground in the OOD set, it can provide additional backgrounds to enhance the effectiveness of the object detector. This might be the reason for the slight improvement of our method from ID+MIX to ID+MIX+OOD.

\input{tabs/tab1}
\input{tabs/tab2}

\textbf{Comparsion on the large-scale dataset.}
Moreover, we show the effectiveness of our method on the large-scale data combination of MS-COCO and OpenImagesv5. We apply DINO pre-trained weight in this experiment following the offline OSSOD, while we use ImageNet pre-trained weight in other experiments.
The result in Table~\ref{tab:openimage} shows that our method can also significantly improve the performance and achieve state-of-the-art under this challenging task.

\begin{table}[t]
\begin{center}
\begin{tabular}{cccc}
   \toprule
   Method & Labeled & Unlabeled & mAP $\uparrow$ \\
   \midrule
   Fully-supervised & COCO & None & 40.90 \\
   \midrule
   UT & COCO & OpenImagesv5 & 41.81 \\
   offline OSSOD & COCO & OpenImagesv5 & 43.14 \\
   Ours & COCO & OpenImagesv5 & \textbf{44.13}     \\
   \bottomrule
\end{tabular}
\end{center}
\caption{Experimental results on COCO-OpenImages.}
\label{tab:openimage}
\end{table}

\begin{table}[t]
\begin{center}
\begin{tabular}{cccccc}
   \toprule
   OOD data & Methods & mAP$\uparrow$ & AUROC$\uparrow$ \\
   \midrule
   \multirow{8}{*}{\makecell{labeled \\ only}} & Energy  & 21.00 & 79.47  \\
    & OE  & 22.55 & 71.99  \\
    & OVA-Net & 23.65 & 78.61  \\
    & VOS & 21.20 & 72.72  \\ 
    & OpenDet & 20.53 & 67.67   \\
    & \cellcolor{black!10}positive head (ours) & \cellcolor{black!10}23.86 & \cellcolor{black!10}72.84   \\
    & offline OSSOD  & 24.40 & 76.26  \\
    \midrule
    \multirow{3}{*}{\makecell{labeled \\ \& \\ unlabeled}} & \cellcolor{black!10}positive head (ours) & \cellcolor{black!10}25.01 & \cellcolor{black!10}77.40   \\
    & \cellcolor{black!10}negative head (ours) & \cellcolor{black!10}25.32          &    \cellcolor{black!10}75.83     \\
    & \cellcolor{black!10}DCO head (ours) & \cellcolor{black!10}\textbf{25.70} & \cellcolor{black!10}\textbf{80.06}   \\
   \bottomrule
\end{tabular}
\end{center}
\caption{Experimental results on different OOD detection methods on COCO-Open with 40 ID classes, 4000 labeled images.}
\label{tab:ablation_ood}
\end{table}

\subsection{Ablation studies and analysis}

\textbf{Ablation study of semi-supervised outlier filtering.} 
We show the benefit of mining more instances from unlabeled data by semi-supervised outlier filtering in Table~\ref{tab:ablation_ood}. The performance of our positive head trained with labeled data only (23.86 mAP) is compared with that trained using both labeled and unlabeled data (25.01 mAP). Note that the positive head is actually of the same structure as the head in the offline OSSOD. We also apply the same OOD score and threshold with offline OSSOD when using positive head only. We also find that applying previous OOD detection-related methods results in relatively lower performance, which aligns with the conclusions drawn in offline OSSOD. This may be because these methods are designed to be trained with abundant labeled data, thus, they are unsuitable for the OSSOD task with limited labeled data. For evaluating these methods, we either utilize officially provided values or employ their value-finding methods to set the thresholds if needed. we also analyze that a higher AUROC does not always ensure a better detection performance, as undetected ID objects(false negative) are not reflected in the AUROC. As a result, using the Energy score gains only 21.00 mAP with 79.47 AUROC, since most of its detection results are false positives with high OOD confidence scores.

\textbf{Effectiveness of the DCO head.} While our method outperforms the previous method with only the positive head using semi-supervised outlier filtering, we find that incorporating our proposed DCO head can further enhance performance. As shown in the last three columns in Table~\ref{tab:ablation_ood}, applying the entire DCO head with both positive and negative heads yields the best performance among all tested methods. We also observe that solely using the negative head results in unstable during the later stages of training.

\textbf{Further analysis of the DCO head.} We further analyze the effectiveness of our DCO head by monitoring the number of ID and OOD pseudo-labels during training. We sample 1000 images from the unlabeled set in COCO-Open with their ID label annotations (these annotations were not used during training). Pseudo-labels having an IoU score over 0.5 with the annotations are considered as ID boxes, otherwise OOD boxes. As shown in Fig.~\ref{fig:coco_id}(b), compared with the positive or negative head only, our DCO head will gradually generate fewer OOD boxes during training but keep a large number of ID boxes. This occurs as the negative head gradually identifies OOD instances with increasing confidence throughout the training process. This phenomenon matches the purpose of designing the DCO head, thus confirming its effectiveness. This experiment is conducted on COCO-Open with 40 ID classes and 4000 labeled images.

\subsection{Additional experiments} 
\begin{table}[t]
\begin{center}
\begin{tabular}{cc|cc}
   \toprule
   Methods & mAP$\uparrow$ & Methods & mAP$\uparrow$ \\
   \midrule
    SoftTeacher & 20.94 & Pseco & 21.90 \\
    SoftTeacher+ours  & \textbf{21.95 \textcolor{RoyalBlue}{(+1.01)}} & Pseco+ours  & \textbf{22.70 \textcolor{RoyalBlue}{(+0.80)}} \\
   \bottomrule
\end{tabular}
\end{center}
\caption{SoftTeacher and Pseco as SSOD frameworks with our method on COCO-Open with 40 ID classes and 4000 labeled images.}
\label{tab:add_ssod}
\end{table}

\begin{table}[t]
\begin{center}
\begin{tabular}{cccc}
   \toprule
   Methods & mAP-COCO$\uparrow$ & mAP$_{50}$-VOC$\uparrow$ \\
   \midrule
    UT    & 28.82 & 81.13      \\
    Ours  & \textbf{30.29 \textcolor{RoyalBlue}{(+1.47)}} & \textbf{81.64 \textcolor{RoyalBlue}{(+0.51)}} \\
   \bottomrule
\end{tabular}
\end{center}
\caption{mAP results on the VOC-COCO benchmark.}
\label{tab:add_data}
\end{table}

\begin{figure}[t]
  \centering
  \includegraphics[width=0.9\linewidth]{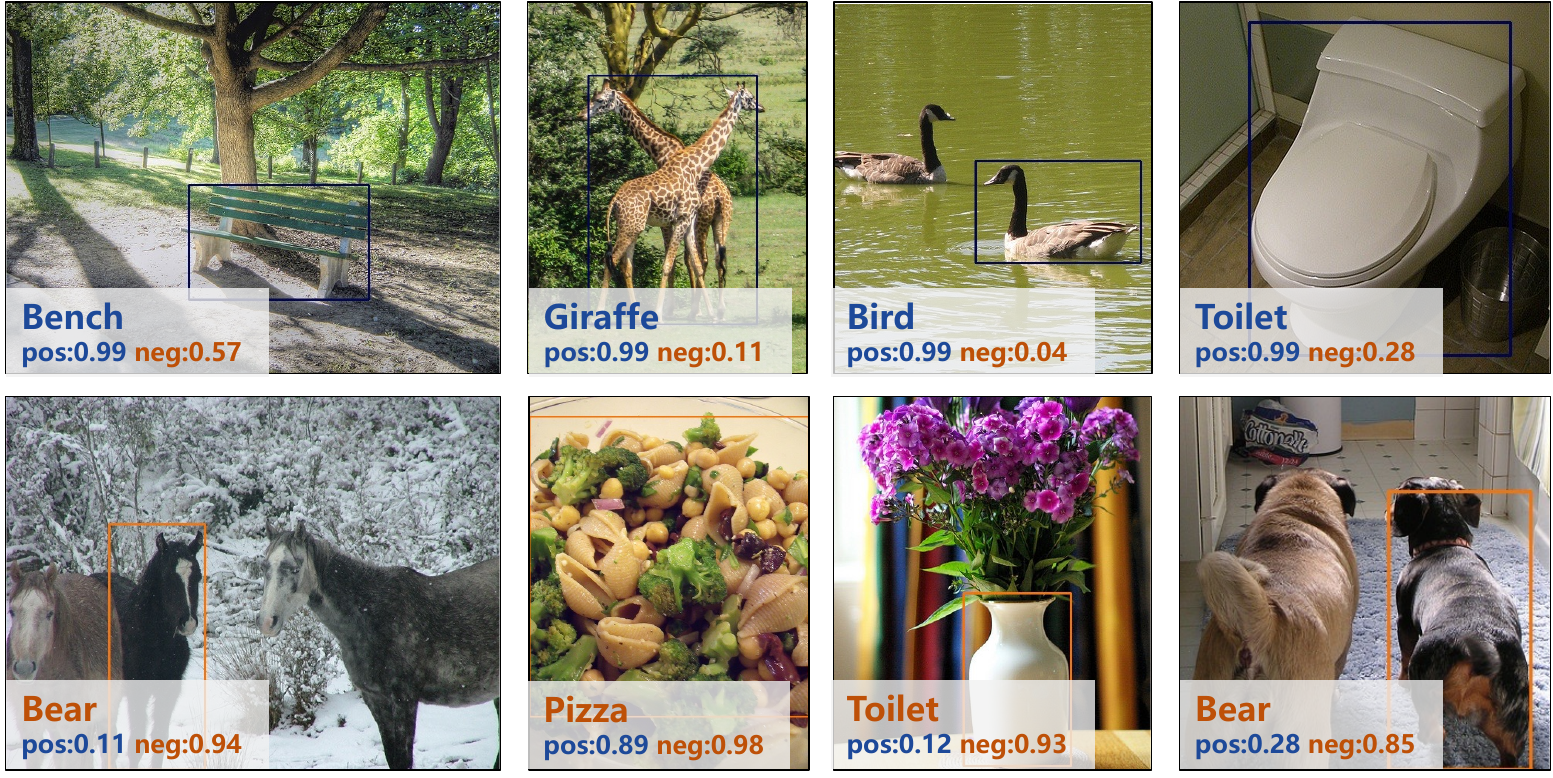}
  \caption{Visualization results of pseudo-labels and related scores from the DCO head (pos: the positive head; neg: the negative head). The instances are predicted as ID (blue) or OOD (orange) by comparing the two scores.}
  \label{fig:add_vis}
\end{figure}

\begin{table}[t]
\begin{center}
\begin{tabular}{ccccc}
\toprule
  & Step & \makecell{Time (iter*s/iter)} & \makecell{GPU Memory (GB)} \\
 \midrule
 \multirow{3}{*}{Offline OSSOD} & 1 & 40k*0.21=8.4k & 50.0 \\
  & 2 & 40k*0.29=11.6k & 57.0 \\
  & 3 & 100k*0.35=35.0k & 59.6 \\
  & Total & 55.0k & 59.6 \\
 \midrule
 Ours & 1 & \makecell{100k*0.34=34.0k \textcolor{RoyalBlue}{(0.62$\times$)}} & 56.5 \\
\bottomrule
\end{tabular}
\end{center}
\caption{Trainging time and GPU memory consumption.}
\label{tab:add_eff}
\end{table}

\textbf{More SSOD frameworks.} We apply our method to two other SSOD frameworks, SoftTeacher~\cite{xu2021end} and Pseco~\cite{li2022pseco}. The results in Table~\ref{tab:add_ssod} show that our method can boost the performance of these SSOD frameworks on the OSSOD task by over 1.0 mAP.

\textbf{More open-set datasets.} Additionally, we evaluate our method on another open-set dataset combination: VOC-COCO. The results in Tab.~\ref{tab:add_data} show that our method can also improve the detection performance on this new benchmark. Note that the Pascal-VOC validation set contains ID instances only, and the MS-COCO validation set contains both ID and OOD instances. Thus, the effectiveness of our method is more significant on the MS-COCO validation set.

\textbf{Efficiency of our method.} Offline OSSOD needs three-step training: 1) train an object detector with labeled data only. 2) train an additional OOD detection network using the proposals from the pre-trained detector. 3) train the SSOD framework with the frozen OOD detection network. Our method only needs to train the entire network once and can converge within the same training iteration. The additional DCO head only consists of two classification heads. Meanwhile, this head can be removed after training. Therefore, our method is more efficient. Table~\ref{tab:add_eff} summarizes the training speed and the GPU memory consumption on the same device of the previous offline method and ours. Our method needs only 0.62$\times$ training time and less memory.

\textbf{Visulization results.} We show some visualization results of ID and OOD pseudo-labels during training in Fig.~\ref{fig:add_vis}. The results are selected from unlabeled data with a detection confidence score over 0.7. Thus they will all be selected as SSOD training instances if there is no OOD filtering. However, the OOD instances in orange boxes will be filtered with our methods. To demonstrate the confidence scores for the positive and negative heads, we only visualized one detection result per image. Actually, there may be other detection results in the image as well. More visualization results are available in the appendix.

\section{Limitations and future direction}

Although we improved the performance by directly removing the detected OOD instances, these instances could potentially serve as useful samples for further training the model, thereby enhancing its detection capabilities. Meanwhile, exploring the distinctions among OOD instances could also be a potential direction, as these instances originally belong to different categories. 

\section{Conclusions}

In this paper, we proposed an online end-to-end trainable OSSOD framework with semi-supervised outlier filtering for utilizing unlabeled data and the Dual Competing OOD head to tackle the error accumulation problem. Experimental results on several benchmarks demonstrate that the proposed method achieves better performance compared with the state-of-the-art OSSOD method. We also conducted ablation studies to validate the effectiveness of each component of our method. And we further show the flexibility of our methods on other SSOD frameworks and open-set datasets. With our proposed method, we can leverage more existing unlabeled data to improve the performance of the model without the need for additional manual filtering OOD instances.

%
%
\bibliographystyle{splncs04}
\bibliography{main}
\end{document}


\title{Online Open-set Semi-supervised Object Detection with Dual Competing Head \\ Supplementary Material} 

\titlerunning{Abbreviated paper title}

\author{First Author\inst{1}\orcidlink{0000-1111-2222-3333} \and
Second Author\inst{2,3}\orcidlink{1111-2222-3333-4444} \and
Third Author\inst{3}\orcidlink{2222--3333-4444-5555}}

\authorrunning{F.~Author et al.}

\institute{Princeton University, Princeton NJ 08544, USA \and
Springer Heidelberg, Tiergartenstr.~17, 69121 Heidelberg, Germany
\email{lncs@springer.com}\\
\url{http://www.springer.com/gp/computer-science/lncs} \and
ABC Institute, Rupert-Karls-University Heidelberg, Heidelberg, Germany\\
\email{\{abc,lncs\}@uni-heidelberg.de}}

\maketitle

\begin{figure*}[h]
  \centering
  \includegraphics[width=0.99\linewidth]{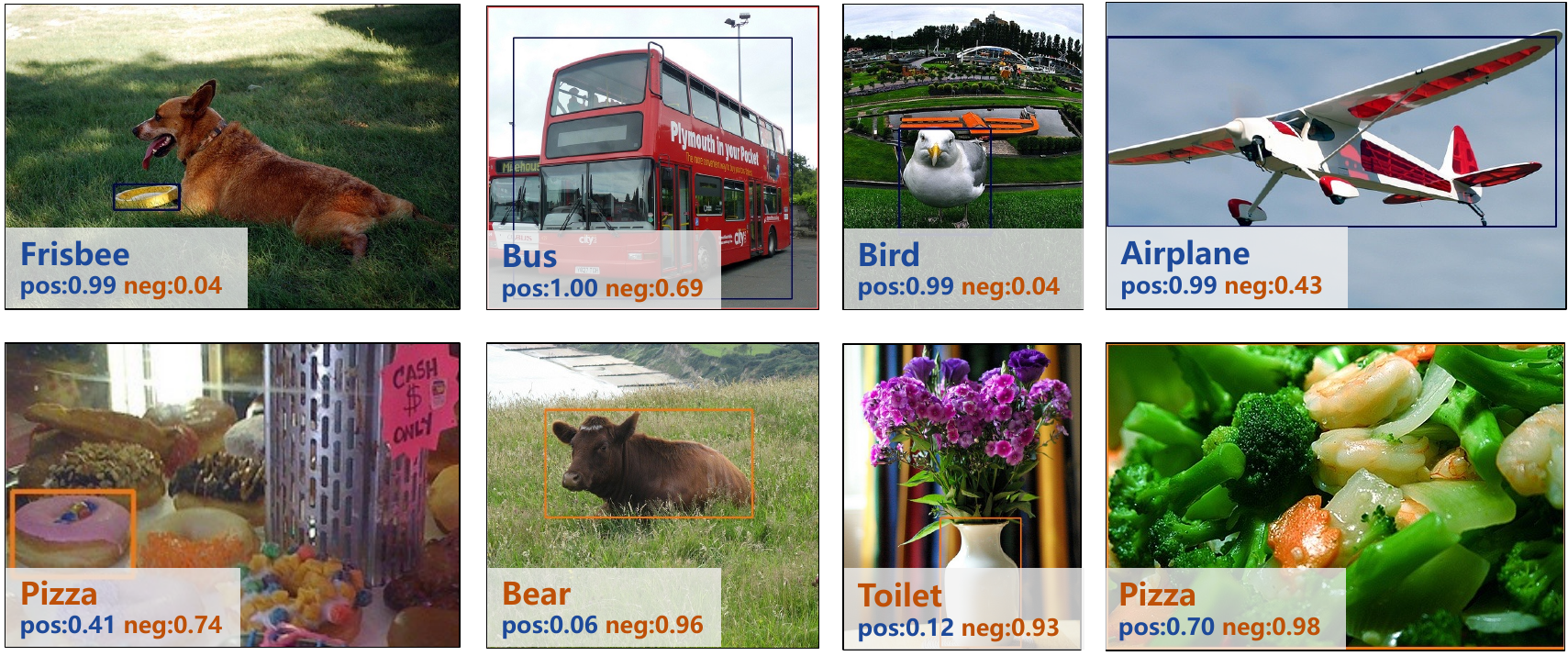}
  \caption{More visualization results of pseudo-labels and related scores from the DCO head (pos: the positive head; neg: the negative head). The instances are predicted as ID (blue) or OOD (orange) by comparing the two scores.}
  \label{fig:compare}
\end{figure*}

We provide a detailed description of our implementation and show more visualization results. Our code is also available in the supplementary material. 


\section{Implementation Details}

\textbf{Dataset.} We randomly sample 20/40/60 ID classes from MS-COCO~\cite{lin2014microsoft} 2017 for COCO-Open. Table~\ref{tab:ablation_classname} lists the selected ID class names for each split.

\noindent\textbf{Environment.} All experiments are conducted with NVIDIA A100 GPU. We use four GPUs for COCO-Open and COCO-VOC, eight GPUs for COCO-OpenImages. We use Python 3.6, PyTorch 1.10.0~\cite{paszke2019pytorch} with CUDA 11.1.

\noindent\textbf{Hyper-parameters.} Our method follows the hyper-parameter settings of our baseline SSOD framework Unbiased Teacher~\cite{liuunbiased}, which is listed in Table~\ref{tab:hypers}. When applying our method on SoftTeacher~\cite{xu2021end} and Pseco~\cite{li2022pseco} in additional experiments, we also follow their settings.

\noindent\textbf{Training.}  We use a batch size of eight labeled and eight unlabeled images for COCO-Open and COCO-VOC, and a batch size of 32 labeled and 32 unlabeled images for COCO-OpenImages. We use the SGD optimizer with a momentum rate of 0.9, a learning rate of 0.01, and a constant learning rate scheduler. Following the Unbiased Teacher~\cite{liuunbiased}, we first introduce a burn-in stage training with the labeled data only. For the COCO-Open and COCO-VOC, we train 180k iterations, which includes 2,000 iterations in the burn-in stage and the remaining iterations in the Teacher-Student Mutual Learning stage. For the COCO-Openimage, we train 720k iterations, which includes 90k iterations in the burn-in stage and the remaining iterations in the Teacher-Student Mutual Learning stage. For the additional experiment of applying our method on SoftTeacher and Pseco, we also follow their training process.

\begin{table*}[t]
\begin{center}
\begin{tabular}{cc}
   \toprule
   No. of ID classes & ID classes \\
   \midrule
   20  & \makecell[l]{\textit{person}, \textit{bicycle}, \textit{bus}, \textit{bird}, \textit{cat}, \textit{horse},  \textit{cow}, \textit{bear}, \textit{backpack}, \textit{handbag}, \\ \textit{skateboard}, \textit{surfboard}, \textit{wine glass}, \textit{fork}, \textit{banana}, \textit{apple}, \textit{orange},  \textit{couch}, \\ \textit{cell phone}, \textit{book}} \\
      \midrule
   40 & \makecell[l]{\textit{person}, \textit{airplane}, \textit{truck}, \textit{boat}, \textit{traffic light}, \textit{fire hydrant}, \textit{stop sign}, \\ \textit{parking meter}, \textit{bench}, \textit{bird}, \textit{cat}, \textit{sheep}, \textit{elephant}, \textit{bear}, \textit{zebra}, \textit{giraffe}, \\ \textit{backpack}, \textit{umbrella}, \textit{handbag}, \textit{frisbee}, \textit{snowboard}, \textit{kite}, \textit{baseball glove}, \\ \textit{skateboard}, \textit{surfboard}, \textit{tennis racket}, \textit{fork}, \textit{knife}, \textit{banana}, \textit{pizza}, \textit{cake}, \\ \textit{couch}, \textit{bed}, \textit{dining table}, \textit{toilet}, \textit{keyboard}, \textit{oven}, \textit{book}, \textit{clock}, \textit{scissors}} \\
   \midrule
    60 & \makecell[l]{\textit{person}, \textit{bicycle}, \textit{car}, \textit{motorcycle}, \textit{bus}, \textit{train}, \textit{truck}, \textit{boat}, \textit{stop sign}, \\ \textit{parking meter}, \textit{dog}, \textit{horse}, \textit{sheep}, \textit{cow}, \textit{bear}, \textit{backpack}, \textit{umbrella}, \\ \textit{tie}, \textit{suitcase}, \textit{skis}, \textit{snowboard}, \textit{sports ball}, \textit{kite}, \textit{baseball bat}, \\ \textit{baseball glove}, \textit{skateboard}, \textit{surfboard}, \textit{bottle}, \textit{wine glass}, \textit{cup}, \\ \textit{fork}, \textit{knife}, \textit{spoon}, \textit{bowl}, \textit{banana}, \textit{apple}, \textit{sandwich}, \textit{orange}, \textit{broccoli}, \\ \textit{carrot}, \textit{hot dog}, \textit{donut}, \textit{chair}, \textit{couch}, \textit{potted plant}, \textit{tv}, \textit{laptop}, \textit{mouse}, \\ \textit{remote}, \textit{cell phone}, \textit{microwave}, \textit{oven}, \textit{toaster}, \textit{sink}, \textit{refrigerator},  \\ 
    \textit{clock}, \textit{vase}, \textit{teddy bear}, \textit{hair drier}, \textit{toothbrush}} \\
   \bottomrule
\end{tabular}
\end{center}
\caption{The ID class names in our COCO-Open dataset.}
\label{tab:ablation_classname}
\end{table*}

\begin{table*}[t]
\begin{center}
\resizebox{\linewidth}{!}{
\begin{tabular}{cccc}
   \toprule
   Hyper-parameter & Description & COCO-Open/VOC & COCO-Openimage \\
   \midrule
    $\tau$ &  Confidence threshold for SSOD &  0.7  & 0.5  \\
   $\lambda$ &  Unsupervised loss weight for SSOD &  4  & 2 \\
   $\alpha$   &  EMA rate &  0.9996  &  0.9996 \\
   $\lambda^{ood}$ & loss weight for the DCO head & 0.1 & 0.1 \\
   \bottomrule
\end{tabular}
}
\end{center}
\caption{Hyper-parameters for our experiments. Only $\lambda^{ood}$ is newly introduced by us.}
\label{tab:hypers}
\end{table*}

\section{More visualization results}

We present some more visualization results of generated pseudo-labels in Figure~\ref{fig:compare} to show the effectiveness of our method. The model is trained on 40 ID classes with 4000 labeled images.
The detected results are selected from unlabeled data with a detection confidence score over 0.7. Thus they will all be selected as SSOD training instances if there is no OOD filtering. However, the instances in orange boxes will be filtered with our methods. To demonstrate the confidence scores for the positive and negative heads, we only visualized one detection result per image. Actually, there may be other detection results in the image as well. These results demonstrate that our method can effectively filter OODs during SSOD, thus leading to better performance.

%
%
\bibliographystyle{splncs04}
\bibliography{main}

%% file: tabs/tab1.tex
\begin{table*}[t]
\centering
\begin{center}
\resizebox{\linewidth}{!}{
\begin{tabular}{cccc}
   \toprule
   \# of ID classes & 20 & 40 & 60 \\
   \midrule
   UT            &  19.06($\pm$0.6) & 21.52($\pm$1.0) &  20.55($\pm$0.5) \\
   offline OSSOD &  19.45($\pm$0.4) \textcolor{RoyalBlue}{(+0.39)} & 24.07($\pm$1.1) \textcolor{RoyalBlue}{(+2.55)} &  22.40($\pm$0.1) \textcolor{RoyalBlue}{(+1.85)} \\
   Ours          &  \textbf{21.09($\pm$0.1) \textcolor{RoyalBlue}{(+2.03)}}  & \textbf{25.57($\pm$0.3) \textcolor{RoyalBlue}{(+4.05)}} &  \textbf{22.47($\pm$0.3) \textcolor{RoyalBlue}{(+1.92)}} \\
   \bottomrule
\end{tabular}
}
\end{center}
\caption{mAP results on COCO-Open under different ID class numbers with 4000 labeled images.}
\label{tab:coco_class}
\end{table*}

%% file: tabs/tab2.tex
\begin{table*}[t]
\centering
\begin{center}
\resizebox{\linewidth}{!}{
\begin{tabular}{cccc}
   \toprule
   \# of labeled imgs & 1000 & 2000 & 4000 \\
   \midrule
   UT            &  13.92($\pm$0.7)  & 15.70($\pm$0.7) & 19.06($\pm$0.6)  \\
   offline OSSOD &  14.47($\pm$1.0) \textcolor{RoyalBlue}{(+0.55)}  & 17.06($\pm$0.2) \textcolor{RoyalBlue}{(+1.36)} & 19.83($\pm$0.3) \textcolor{RoyalBlue}{(+0.77)}  \\
   Ours          &  \textbf{15.70($\pm$0.8) \textcolor{RoyalBlue}{(+1.78)}} & \textbf{18.54($\pm$0.7) \textcolor{RoyalBlue}{(+2.84)}} & \textbf{21.09($\pm$0.1) \textcolor{RoyalBlue}{(+2.03)}} \\
   \bottomrule
\end{tabular}
}
\end{center}
\caption{mAP results on COCO-Open under different labeled image numbers with 20 ID classes.}
\label{tab:coco_label}
\end{table*}